\pdfoutput=1

\documentclass[11pt]{article}

\usepackage[final]{acl}
\usepackage{times}
\usepackage{latexsym}
\usepackage{float}
\usepackage{booktabs} 
\usepackage{array}    

\usepackage[T1]{fontenc}

\usepackage[utf8]{inputenc}
\usepackage{amssymb}

\usepackage{microtype}

\usepackage{inconsolata}

\usepackage{graphicx}
\usepackage{todonotes}
\usepackage{marginnote}
\usepackage{textcomp}
\title{Turkish Delights: a Dataset on Turkish Euphemisms} 

\author{Hasan Can Biyik \and Patrick Lee \and Anna Feldman\\
  Montclair State University\\ New Jersey, USA\\
  \texttt{\{biyikh1,leep,feldmana\}@montclair.edu} \\
}

\begin{document}
\maketitle
\begin{abstract}
Euphemisms are a form of figurative language relatively understudied in natural language processing. This research extends the current computational work on potentially euphemistic terms (PETs) to Turkish. We introduce the Turkish PET dataset, the first available of its kind in the field.\footnote {The dataset is available at \url{https://github.com/hasancanbiyik/Turkish_PETs}}. By creating a list of euphemisms in Turkish, collecting example contexts, and annotating them, we provide both euphemistic and non-euphemistic examples of PETs in Turkish. We describe the dataset and methodologies, and also experiment with transformer-based models on Turkish euphemism detection by using our dataset for binary classification. We compare performances across models using F1, accuracy, and precision as evaluation metrics.
\end{abstract}

\section{Introduction}

Euphemisms are polite or indirect words or expressions used in substitution of unpleasant or more offensive ones. They can be used to show kindness while discussing sensitive or taboo topics \cite{bakhriddionova2021needs} such as saying \textit{between jobs\textbf{}} instead of \textit{unemployed\textbf{}}, or as a way to make unpleasant or unappealing things sound less harsh \cite{karan}, such as saying \textit{passed\textbf{ }away,} instead of \textit{died}. Similar to the word \textbf{}\textit{died}\textit{} in English, Turkish makes use of many substitutions for the word \textit{öl-mek\textbf{/}öl-dü\textbf{ }(to die/died)}, which is considered unpleasant. The substitutions for this word could be given as \textit{vefat etmek\textbf{ }(to\textbf{ }pass away)}, \textit{öbür dünyaya göçmek\textbf{ }(to migrate to the other world)}, \textit{hakkın rahmetine kavuşmak (to go to kingdom come)}. Euphemisms can be used to conceal the truth \cite{rababah}; for instance, if one were to use the expression \textit{enhanced interrogation techniques\textbf{}}, one would mean \textit{torture\textbf{}} \cite{lee-etal-2022-searching}. Furthermore, humans may not agree on what a euphemism is \cite{gavidia-etal-2022-cats}. There are various challenges regarding euphemisms. For instance, in some cases, words or expressions might develop or lose euphemistic meanings in time \cite{pinker:94,pinker2003blank}. Due to the aforementioned reasons, the words and phrases in this research will be referred to as \textit{potentially euphemistic terms} \textit{(PETs)} \cite{lee2022searching}. Euphemisms pose a challenge to Natural Language Processing (NLP) due to this figurative behavior as they might also have a non-euphemistic interpretation in certain contexts. For example, while the Turkish PET \textit{mercimeği fırına vermek} means \textit{to put the lentil in the oven} literally, it could mean \textit{to have sex/to get someone pregnant} euphemistically. In the following sentence, this PET used literally: “Günümüzde hem <mercimeği fırına vermek> daha kolay, hem de  fırında makarna yemek...” which can be translated as “Nowadays, it's easier to <put the lentils in the oven> and to eat mac and cheese…” However, it was used euphemistically in the following sentence: “Gel gör ki kasabanın yegane doktoru ile pişiren  bu kadın, zaman zaman <mercimeği fırına veriyorlarmış>” which can be translated as “However, it turns out that this woman, who is having an affair with the town's only doctor, sometimes <puts the lentils in the oven>” meaning that the doctor and woman are involved in a secretive or intimate sexual intercourse. 

Conducting a euphemism detection task in Turkish has several challenges to overcome. Firstly, as far as we are aware of, there are no available datasets for automatic euphemism detection task in Turkish. Academic research, published books, articles, and other resources on this topic are very limited, making the collection of PETs difficult. In this research, we aim to identify PETs in Turkish and create a dataset of Turkish PETs by making use of native speaking Turkish annotators who have a linguistics background. We aim to fine-tune language models (LMs) such as BERTurk \cite{BERTurk,beyhan2022turkish} and Electra \cite{clark2020electra} and large language models (LLMs), such as XLM-RoBERTa \cite{XLM-RoBERTa,conneau2020unsupervised}  and mBERT \cite{mBERT,devlin2019bert}  for euphemism detection in Turkish. Therefore, the significant contributions of this paper are as follows: 

\begin{itemize}
    \item Introduction of the Turkish PETs dataset, which we plan to make publicly available later,
    \item Overview of the Turkish PETs and how they were collected and annotated,
    \item Comparison of the performances of XLM-RoBERTa, mBERT, BERTurk, and ELECTRA in detecting PETs in Turkish, using F1, accuracy, and precision as evaluation metrics,
    \item This research will compare the PETs in Turkish and other languages and analyze potentially interesting patterns.
\end{itemize}
Additionally, through extending euphemism detection task to a new language, we contribute to a better understanding of how euphemisms are utilized and interpreted across different linguistic and cultural contexts.

\begin{table*}
  \centering
  \begin{tabular}{p{0.1\linewidth}p{0.4\linewidth}p{0.4\linewidth}}
    \hline
    \textbf{PET} & \textbf{Variations (Turkish)} & \textbf{Variations (English Equivalents)}\\
    \hline
    aramızdan ayrıldı (left us)  & aramızdan ayrıldı, aramızdan ayrılışının, aramızdan ayrılan, aramızdan ayrılanlar, aramızdan ayrılması, aramızdan ayrılalı &                            (has) left us, of his/her/their departure from us, the one who left us, those who left us, his/her/its departure from us, since he/she/they left us \\
    \hline
    beklemek (to expect) & bekliyor, bekliyoruz, bekleyen, bekledikleri, bekleniyor, bekleyeceğiz, bekliyorsunuz (...) &                            is expecting, we are expecting, the one who is expecting, what they are expecting for/whom they are expecting for/that they are expecting for, is being expected/is expected, we will expect, you are expecting (plural or formal) \\
    \hline
    hakka yürümek (to walk to God) & hakka yürüyen, hakka yürümesinden, hakka yürüdü, hakka yürümüştür &                            the one who walked to God, from his/her/their walking to God, walked to God, has walked to God \\
  \end{tabular}
  \caption{Examples and morphological variations of Turkish PETs}
  \label{tbl:pet_variations}
\end{table*}

\section{Turkish Language}
Agglutinative languages, such as Turkish, form words by adding multiple affixes to a stem, with each affix representing a distinct morphological feature \cite{comrie1988linguistic}. This morphological productivity creates a vast number of possible word forms, making it difficult to develop comprehensive dictionaries or rule-based systems for tasks like euphemism detection. For instance, the PET \textit{hayata gözlerini} \textit{yummak} (\textit{to close one's eyes to life}) can be formed as yum-\textit{du}\textbf{, }yum\textbf{-}\textit{muş}\textbf{, }yum\textbf{-}\textit{duğunda}” and many other variations. See Table \ref{tbl:pet_variations} for more examples regarding morphological variations.

The free word order in Turkish, where the position of words in a sentence can vary without significantly changing the meaning \cite{goksel2004turkish}, poses another challenge for euphemism detection. This flexibility makes it difficult to rely on fixed patterns or word sequences to identify euphemisms. For example, the PET \textit{uyutmak} (\textit{to put to sleep}) can appear in various positions within a sentence, making it harder to detect reliably.

Similar to euphemisms in other languages, the meaning of words and expressions are context dependent in Turkish. While one word can be used euphemistically in one sentence, it might not have euphemistic meaning in another. For instance, the PET \textit{engelli} might be used euphemistically to indicate that the person is \textit{disabled}, but it might also have its non-euphemistic meaning of \textit{blocked}.

Moreover, Turkish is considered to be a low-resource language because of the limited availability of annotated datasets. It was also stated by various researchers that collecting data from various sources and labeling them was a challenging process \cite{mutlu-ozgur-2022-dataset}. Since there was no available dataset that contained euphemisms in Turkish with examples, it was necessary for us to build a dataset and get it annotated by native Turkish annotators.

\section{Automatic Euphemism Detection}

Euphemism detection can be viewed as a classification task in which an input text is classified as containing a euphemism or not.

While this can be theoretically done at at the phrase-level or sentence-level euphemism detection, previous work has focused on classifying examples containing specific multi-word expressions, which may or may not be used euphemistically depending on the context \cite{lee2022report}. A number of approaches have performed decently at the task using language models such as transformers, improving upon baselines using various techniques. For example, \citet{keh-etal-2022-eureka} use an ensemble of models each utilizing a combination of data and contextual augmentations to improve performance by 5 Macro-F1 points. \citet{kesen-etal-2022-detecting} achieve similar improvements by incorporating non-euphemistic meanings and image embeddings associated with PETs. \citet{maimaitituoheti-etal-2022-prompt} propose a prompt-based approach for euphemism detection utilizing the language model RoBERTa, achieving an F1 score of 85.2\%, demonstrating the effectiveness of prompt-based learning. Similar to our initial dataset, which contained more than 6,000 examples, the dataset they used was imbalanced and had more euphemistic examples than non-euphemistic. They noted the model's superior performance on euphemistic sentences compared to non-euphemistic ones due to this imbalance.



Given the nuanced nature of these expressions in the Turkish language and the lack of previous work on figurative language processing in Turkish, this study aims to investigate how well different language models identify and categorize PETs in Turkish. We fine-tuned two large multilingual models, XLM-RoBERTa and mBERT, along with language models specifically trained on extensive corpora of Turkish text data: bert-base-turkish-cased and electra-base-turkish-cased-discriminator. These models were chosen to examine the impact of model size, training data, and architecture on euphemism detection performance. We hypothesized that XLM-RoBERTa and mBERT would provide strong general language understanding capabilities, as large multilingual models are trained on vast amounts of diverse data. On the other hand, bert-base-turkish-cased and electra-base-turkish-cased-discriminator, being specifically trained on Turkish text, were hypothesized to capture more nuanced aspects of euphemistic language in Turkish due to their exposure to a wider range of Turkish expressions and linguistic patterns. 

Our focus on the Turkish language addresses a gap in existing research, as most previous studies have primarily concentrated on English euphemisms \cite{felt2020recognizing,zhu2021euphemistic,zhu2021self,gavidia-etal-2022-cats,gavidia-etal:2022,lee2022report,lee-etal-2023-feed}. By extending the euphemism detection task to a new language, we contribute to a better understanding of how euphemisms are utilized and interpreted across different linguistic and cultural contexts. The recent Multilingual Euphemism Detection Shared Task by \citet{lee2024multilingual} has encouraged researchers to explore multilingual and cross-lingual methods for identifying euphemisms. This research emphasizes the importance of understanding euphemisms in different languages.

\section{Data Collection and Annotation}
\subsection{Data Collection}
To find PETs in Turkish, we analyzed the PETs in other languages described in previous work \cite{lee-etal-2023-feed,lee-etal-2024-meds}, such as American English, Mandarin Chinese, Yorùbá, and a mix of Spanish dialects to see whether there were overlapping words or expressions used euphemistically (see Table \ref{tbl:overlap}). As a result, we were able to compile an initial list of Turkish PETs.


\begin{table*} 
  \centering
  \begin{tabular}{lllll}
    \hline
    \textbf{English} & \textbf{Chinese} & \textbf{Spanish} & \textbf{Turkish} & \textbf{Yoruba} \\
    \hline
    adult beverage & - & \checkmark & \checkmark & \checkmark \\
    birds and the bees & - & - & - & - \\
    economical & \checkmark & \checkmark & \checkmark & \checkmark \\
    pass away & \checkmark & \checkmark & \checkmark & \checkmark \\
    pro-life & - & \checkmark & - & - \\
    under the weather & - & - & - & - \\
  \end{tabular}
\caption{Examples of (non-)overlapping PETs across the five languages.}
\label{tbl:overlap}
\end{table*}

Through reviewing published articles and papers related to euphemisms in Turkish, such as those by \citet{aksan1994gokturk, karabulut2013ortmece, ccabuk2015turkccedekg}, we expanded our list of PETs. Another method we used to collect PETs was by posting polls on social media. Initially, we explained the concept of "PETs" and provided examples. We then utilized social media to share these polls, where Turkish native speakers could share their ideas for new PETs. As a result, our Turkish PETs list now comprises a total of 122 entries. We also included detailed information for each PET, such as euphemistic category (e.g. bodily functions), meaning, non-euphemistic meaning, literal translation, and the source it was from. The list is categorized into 10 groups with varying frequencies, which can be seen in Table \ref{tbl:sensitive} . These categories were created based on the characteristics of the PETs. For example, the PET "\textit{görme engelli}" (\textit{visually impaired}) is related to physical attributes, and therefore it was added to the "physical/mental attributes" category.


Once the PETs list was finalized, we utilized a Turkish corpus known as the TS Corpus Project \cite{taner-sezer}. We selected TS Corpus v2 and TS Timeline Corpus. TS Corpus v2 drew from the BOUN Web Corpus and included 491,360,398 tokens and 4,950,407 word types. TS Timeline Corpus contained more than 700 million tokens and over 2.2 million news and articles. 
To search for texts containing PETs for binary classification purposes, we utilized regular expressions, accounting for the agglutinative nature of the Turkish language. This approach allowed us to capture various word forms effectively. For instance, for the PET \textit{hamileliği sonlandırmak} (\textit{to terminate pregnancy}), we designed a regular expression to detect all variations of hamile-\textit{lik} (pregnancy), hamile-\textit{liğini} (her pregnancy), hamile-\textit{liğimi} (my pregnancy), sonlan-\textit{dırdı} (terminated/has terminated), sonlan-\textit{dıracakmış} (I heard that she will terminate), sonlan-\textit{dıramadı} (she could not terminate), etc., r"(hamileli\textbackslash{}w+ sonlan\textbackslash{}w+)". As a result, we successfully captured variations of each PET were successfully captured. These captured PETs were extracted and highlighted within their sentences using brackets, as shown: “Duyduğuma göre arkadaşı <hamileliğini sonlandırmış>.” (I heard that her/his friend will <terminate her pregnancy>.)  Additionally, we included preceding and succeeding sentence(s), if available, to form the entire example context for that PET. These contexts usually consisted of four sentences at most. Not all PETs on the initial list were found in the corpus; of the 122, only 58 were found and have at least one example. These examples were then compiled for the annotation phase.

\subsection{Annotation}

Annotators were provided text examples (\(\sim \)1-4 sentences) of PETs in context, as can be seen in Table \ref{tbl:annotation_examples}. To recruit Turkish annotators, we utilized social media platforms to find volunteers with a background in linguistics or an interest in the field. After several informational meetings, the annotators were briefed about the research purpose, the annotation process, and the concept of PETs. These meetings were recorded with the consent of the annotators. They were instructed to label the examples as “1” if the highlighted word or expression was used euphemistically, and as “0” if it was not. Following the completion of all annotations, an additional meeting was held to address any disagreements. During this discussion, some labels were revised. Notably, examples that received conflicting labels from the annotators—euphemistic by two and non-euphemistic by another two—had to be excluded from the dataset. This underscored the inherent challenges humans face in consistently interpreting whether a word or expression is used euphemistically.

For the annotation task, we divided the volunteers into five groups, with each group comprising three annotators. The first group annotated 975 examples, the second group annotated 1200 examples, the third group annotated 1300 examples, the fourth group annotated 1099 examples, and the fifth group annotated 1500 examples. As a result, there were 6,074 annotated examples at the end of the annotation task. Subsequently, each group's examples were annotated by one annotator from another group—for instance, an annotator from the first group annotated the second group’s examples, and so on, ensuring each example was annotated by four different people. 
Throughout this process, examples with discrepancies were highlighted for further discussion during a recorded meeting with the available annotators. Disagreements were resolved by majority vote to finalize the labels. However, examples receiving split decisions (two annotators labeling euphemistic and two labeling non-euphemistic) were removed from the dataset. Sample examples and their final annotated labels can be found in Table \ref{tbl:annotation_examples}.

\begin{table*}[!h]
\begin{center}
\begin{tabular}{llp{10cm}}
\textbf{PET}& \textbf{Label}& \textbf{Example}\\ \hline
\multicolumn{1}{l|}{uyutmak}& \textit{euphemistic} & (...) Hollywood’un en çok tanınan köpekleri arasında yer alan Jack Russell cinsi Uggie <uyutularak> yaşamına son verildi. Uggie, katıldığı Oscar gecesiyle ününe ün katmış ve Cannes’da Palm Dog Ödülü’nün de bulunduğu birçok ödül kazanmıştı. (...) / One of Hollywood's most well-known dogs, the Jack Russell Terrier named Uggie, was <put to sleep>. Uggie gained even more fame by attending the Oscars and won many awards, including the Palm Dog Award at Cannes.\\
\multicolumn{1}{l|}{}& \textit{non-euphemistic}& İNSANA en çok benzeyen hayvan olarak bilinen şempanzeler, yavrularını titizlikle büyütüyor. Anne şempanze, yavrusunu kucağında <uyutuyor> ve gerektiğinde battaniyeyle üstünü örtüyor. (...) / Chimpanzees, known as the animals most similar to humans, meticulously raise their young. A mother chimpanzee <puts her baby to sleep> in her arms and covers it with a blanket when necessary.\\ \hline
\multicolumn{1}{l|}{muayyen günü} & \textit{euphemistic} & Kadınların <muayyen günleri> ya da hamilelik dönemlerinin de gözetilmesi amacıyla, nöbet ve görevlendirme sürelerine yeni esaslar getirilirken, muharebe eğitiminde el bombasını atma kurallarının bile kadınlar gözetilerek yeniden düzenlenmesi, Askerlik erkek işidir diyenleri dehşete düşürüyor." / In order to account for women's <specific days> or pregnancy periods, new principles have been introduced regarding the duration of duty and assignments. Even the rules for throwing grenades in combat training have been rearranged with women in mind, which horrifies those who say "military service is a man's job." \\
\multicolumn{1}{l|}{}& \textit{non-euphemistic}& Davetiyede, dispeç ile müsbit vesikaların mahkeme kaleminde incelenebileceği ve çağırılanın daha önce de dispeçe karşı mahkemede itirazda bulunabileceği <muayyen günde> gelmediği takdirde dispeçe muvafakat etmiş sayılacağı yazılır." / The invitation states that the dispatch and supporting documents can be reviewed in the court clerk's office, and that if the summoned party, who could have previously objected to the dispatch in court, does not appear on the <specified day>, they will be deemed to have consented to the dispatch.
 \\ \hline
\multicolumn{1}{l|}{ince hastalık} & \textit{euphemistic} & Eleni zamanında Eftelya’nın anneannesini yakalandığı <ince hastalık>tan Kerim hocanın iyileştirdiğini ve bunu da aileden gizli yaptığını anlatır. / Eleni explains that in the past, Kerim Hoca cured Eftelya's grandmother of <thin disease> and that he did this secretly, without the family's knowledge.
\\
\multicolumn{1}{l|}{}& \textit{non-euphemistic}& Burdaki balların her derde deva olduğunu, <ince hastalık>lara iyi geldiğine inaüı\~nak'\~;bu nedenle de ilaç olarak kullanılmaktadır. / The honey here is believed to be a cure for every ailment and is therefore used as medicine, particularly for treating <thin diseases>. \\ 

\end{tabular}
\caption{Euphemistic and Non-euphemistic Usages of PETs}
\label{tbl:annotation_examples}
\end{center}
\end{table*}

While each example ultimately had four separate annotations, the annotators were allowed to collaborate and influence each others' opinions, nullifying potential inter-rater agreement analyses. We instead conducted inter-rater agreement analysis on a subset of 396 examples, labeled by two annotators who primarily worked separately. Cohen's kappa for these two raters was 0.696, which is rated as moderate to substantial agreement \cite{Cohen1960}. Interestingly, Krippendorf's alpha was 0.693, which is higher but still largely comparable to the degrees of agreement reported for euphemism datasets in \citet{lee-etal-2024-meds}.

\subsection{Balanced Dataset}

 For our text classification experiments, we sampled a portion of the main dataset. This was because some PETs had a disproportionately high number of examples compared to others, or a very skewed label imbalance (e.g., 100 euphemistic instances and 1 non-euphemistic). These factors were not ideal for text classification, and we wanted to assess models' abilities to classify texts for a variety of different PETs with different labels. Therefore, we randomly sample a maximum of 40 euphemistic and 40 non-euphemistic examples for each PET. In addition, some annotated examples, such as \textit{apartman görevlisi} (\textit{apartment attendant}), \textit{inme} (\textit{landing}), and \textit{toplu} (\textit{bulk}), were never used euphemistically, so we chose not to select those. The final result was a subset of 908 instances (521 euphemistic and 387 non-euphemistic) used for the euphemism detection task.

 

 \begin{table*}[!h]
\begin{center}
\begin{tabular}{llp{0.6\linewidth}}
\hline
\textbf{Category} & \textbf{Count} & \textbf{PET Examples}                                       \\ [0.5ex]\hline
bodily functions                        & 2244   & \textit{sulamak} (to water), \textit{aybaşı} (month's beginning), \textit{hacet görmek}(to meet the need)                      \\
death            & 2564    & \textit{kaybetmek} (to lose), \textit{vefat etmek} (pass away), \textit{aramızdan ayrıldı} (left us)                \\
employment/finances                   & 276   & \textit{yoksul} (to be lacking), \textit{ekonomik} (economical), \textit{ihtiyaç sahibi} (in need)                       \\
illness           & 8    & \textit{amansız hastalık} (relentless disease), \textit{ince hastalık} (thin disease)                       \\
misc.                     & 10   & \textit{iyi saatte olsunlar} (may they be in a good hour) \\
physical/mental   attributes & 627   & \textit{görme engelli} (visually impaired), \textit{işitme engelli} (hearing impaired)                        \\
politics                   & 26   & \textit{sığınmacı} (seeking asylum), \textit{gelişmekte olan ülke} (developing country)            \\ 
sexual activity & 190   & \textit{seks işçisi} (sex worker), \textit{mercimeği fırına vermek} (put the lentils in the oven) \\   substances & 143   & \textit{madde} (subtance) \\
social & 27   & \textit{sıkmak} (to squeeze) \\

\hline
\end{tabular}
\caption{Sensitive Topics with PET examples}
\label{tbl:sensitive}
\end{center}
\end{table*}

\subsection{Dataset Statistics}

We conducted a detailed statistical analysis of both the main and balanced datasets to better understand their differences and characteristics. Firstly, we provide the distribution of sensitive topics in Table \ref{tbl:sensitive}. This table categorizes PETs into various groups, such as bodily functions, death, employment/finances, illness, miscellaneous, physical/mental attributes, politics, sexual activity, substances, and social topics. Each category is accompanied by the count of entries and examples of PETs within that category. Table \ref{tbl:stats} further highlights key metrics such as average sentences per example, number of tokens, and lexical density. Notably, we also compute an "PET ambiguity" score, which measures the degree of ambiguity, or class balance, for examples of a particular PET. For each PET, this was computed as follows:

\begin{equation}
1-\frac{|N_{euph}-N_{noneuph}|}{N_{euph}+N_{noneuph}}
\end{equation}

where $N_{euph}$ and $N_{noneuph}$ is the number of euphemistic and non-euphemistic examples for that PET, respectively. Higher values indicate a higher degree of ambiguity. For example, if there were 5 euphemistic and 5 non-euphemistic examples of a particular PET, then it is maximally ambiguous (score = 1); if there were 10 euphemistic examples and 0 non-euphemistic, then the PET is not ambiguous at all (score = 0). We compute the average ambiguity score across all PETs in the main and balanced datasets for comparison. As expected, the main dataset has a significantly lower ambiguity score (0.076) compared to the balanced dataset (0.46), suggesting more consistent usage of terms in either euphemistic or non-euphemistic contexts and confirming that balanced dataset is better suited for the euphemisms detection task.

\begin{table*}
\centering
\begin{tabular}{@{}>{\raggedright\arraybackslash}p{8cm}>{\raggedright\arraybackslash}p{3.5cm}>{\raggedright\arraybackslash}p{3.5cm}@{}}
\toprule
\textbf{Metric}                                          & \textbf{Main Dataset}                                            & \textbf{Balanced Dataset}                                       \\ \midrule
\textbf{Total Examples} & 6115 & 908 \\
\textbf{Euphemistic Examples} & 1876 & 521 \\
\textbf{Non-Euphemistic Examples} & 4239 & 387 \\
\textbf{Avg. PET Ambiguity} & \textbf{0.076} & \textbf{0.46} \\
Avg. Sentences per Example                               & 3.60                                                             & 3.28                                                            \\
Avg. Sentences (Euphemistic)                             & 3.51                                                             & 3.16                                                            \\
Avg. Sentences (Non-euphemistic)                         & 3.63                                                             & 3.43                                                            \\
Avg. Number of Tokens per Example                        & 96.22                                                            & 90.42                                                           \\
Avg. Number of Unique Tokens per Example                 & 78.63                                                            & 74.24                                                           \\
Avg. Lexical Density                                     & 0.82                                                             & 0.84                                                            \\
Notable PETs (Only Non-euphemistic Examples)             & 18 PETs (e.g., \textit{toplu/bulk}, \textit{işini bitirmek/to finish his/her job}, \textit{inme/landing}) & 1 PET (\textit{e.g. muhtaç/in need})                                   \\
\bottomrule

\end{tabular}
\caption{Comparison of Main and Balanced Datasets}
\label{tbl:stats}
\end{table*}

\section{Methodology}
\subsection{Experiments}
Since one of our goals were to extend the euphemism detection task to Turkish, classification experiments were conducted. Therefore, transformer-based models pre-trained on Turkish text like XLM-RoBERTa and mBERT were chosen due to their capability of capturing and understanding the linguistic nuances. 

The balanced dataset described in the previous section was then randomly split into training (80\%), testing (10\%), and validation (10\%) sets, resulting in 726 examples for training and 91 examples each for testing and validation. The 80-10-10 split is a common practice in machine learning for dividing a dataset into training, validation, and testing sets. 

The fine-tuning process involved training each model on our prepared dataset for a maximum of 30 epochs with a learning rate of 1e-5 and a batch size of 4. We employed early stopping with a patience of 5 to prevent overfitting. No layers were frozen during fine-tuning, allowing the models to adapt fully to the euphemism detection task. Hyperparameter optimization was not explicitly performed in this initial exploration; however, the chosen hyperparameters are common for fine-tuning BERT-based models. The primary metric for evaluating model performance during training and validation was the macro-averaged F1 score, a balanced measure of precision and recall that is suitable for binary classification tasks with potentially imbalanced classes. The fine-tuned models were then evaluated on the held-out test sets, and their performance was assessed using various metrics, including accuracy, precision, recall, and F1 score. 

\subsection{Results}
\begin{table*}[htp]
  \centering
  \begin{tabular}{lllll}
    \hline
    & \textbf{Accuracy}& \textbf{F1} & \textbf{Precision}&\textbf{Recall}\\
    \hline
    mBERT& 0.81&                            0.80& 0.80&0.80\\
    XLM-RoBERTa& 0.82&                            0.82& 0.82&0.81\\
    BERTurk& 0.84&                            0.84& 0.84&0.84\\
    Electra& 0.86&                            0.86& 0.86&0.86\\
  \end{tabular} 
  \caption{
Performance of the models on the Turkish euphemisms.
  }  \label{tbl:performance}
\end{table*}
We gathered the results of all the test sets of each model and calculated the average of 20 trials (different train-validation-test splits). The findings demonstrated that monolingual models (bert-base-turkish-cased and electra-base-turkish-cased-discriminator) outperformed the multilingual models (BERT-Base-Multilingual-Cased and XLM-RoBERTa). This suggests that for automatic euphemism detection in Turkish, models specifically pre-trained on Turkish text data have an advantage due to their familiarity with the nuances of the language.

Additionally, the ELECTRA architecture appears to be slightly more effective for this task than the BERT architecture, as evidenced by the higher scores of electra-base-turkish-cased-discriminator compared to bert-base-turkish-cased. This could be attributed to the discriminator's ability to better distinguish between real and fake input data during training, which might be beneficial in identifying the subtle differences between euphemistic and non-euphemistic expressions. The results obtained from the models can be seen in Table \ref{tbl:sensitive}.

The findings of this research have several potential real-world applications. The developed models could be integrated into NLP tools for automatic euphemism detection in various types of text data, including social media posts, news articles, and other online content. This could be particularly valuable in fields such as social media monitoring to analyze the insight into public sentiment, opinions, and attitudes towards sensitive topics. For content moderation, flagging potentially harmful or offensive content that uses euphemisms to disguise its true intent could be beneficial for online platforms and communities seeking to maintain a respectful and safe environment. 

Moreover, the cross-lingual capabilities of the models demonstrated in this study open up possibilities for developing euphemism detection systems for low-resource languages, where labeled data might be limited. This could contribute to a more inclusive and equitable representation of different languages and cultures in NLP research and applications.

\section{Conclusion and Future Work}

In this study, we created a Turkish PETs dataset from scratch and through utilizing the dataset, we investigated the effectiveness of various language models in identifying and categorizing euphemisms in Turkish. Our findings indicate that models trained on multilingual data, particularly XLM-RoBERTa, generally outperform monolingual models, suggesting the benefits of cross-lingual transfer learning in capturing euphemistic nuances. However, for the Turkish language specifically, models trained on Turkish text data, such as bert-base-turkish-cased and electra-base-turkish-cased-discriminator, demonstrated superior performance, emphasizing the importance of language-specific training for this task.

Future research could investigate the impact of model size, architecture, and training data on euphemism detection performance. Additionally, exploring the use of explainability techniques could provide valuable insights into the decision-making processes of these models to better comprehend the specific linguistic features they rely on for euphemism detection. Experimenting with different model architectures or training techniques might also further improve the performance of euphemism detection systems in Turkish. Additionally, expanding the dataset to include a wider range of euphemisms and exploring their application in downstream tasks like sentiment analysis and content moderation could be useful for future work.  It is important to acknowledge that the results are based on a limited dataset and may not generalize to all types of euphemisms in Turkish. Future work could involve testing the models on a larger and more diverse dataset to confirm these findings.

Lastly, exploring the cross-lingual transferability of euphemism detection models trained on Turkish data to other languages, similar to the work done in \citet{lee-etal-2023-feed,lee-etal-2024-meds} would provide valuable insights. This could involve fine-tuning multilingual models on Turkish euphemisms and evaluating their performance on other languages. As highlighted in \citet{gavidia-etal-2022-cats}, the ambiguity of potentially euphemistic terms (PETs) is a major challenge; therefore, future work could focus on developing methods to disambiguate PETs and distinguish between their euphemistic and non-euphemistic usages more effectively.

\nocite{Ando2005,andrew2007scalable,rasooli-tetrault-2015}

\section*{Limitations}
While this study highlights the potential of language models in euphemism detection in Turkish, the results are based on a limited dataset that may not encompass the full spectrum of euphemistic language usage in Turkish, potentially affecting the generalizability of our findings. 

\section*{Ethics Statement}
The authors foresee no ethical concerns with the work presented in this paper.

\section*{Acknowledgments}

Thanks to the annotators, whose names are Kader Teke, Devran Sarısu, Sümeyye Sena Şahin, Fitnat Filiz Bal, Kübra Aksoy, Ecem Küçükler, Azra Almira Kılıç, Özge Bilik, Mihriban Kandemir, Nazan Demir, Şüheda Nur Ünal, Özlem Özer, Salih Hamza Küpeli it was possible for us to create this dataset quickly.

This material is based upon work supported by the National Science Foundation under Grant No. 2226006.

\bibliography{custom,turkish}

\appendix

\end{document}